\documentclass{article}

\usepackage[preprint]{neurips_2021}

\usepackage[utf8]{inputenc} 
\usepackage[T1]{fontenc}    
\usepackage{hyperref}       
\usepackage{url}            
\usepackage{booktabs}       
\usepackage{amsfonts}       
\usepackage{nicefrac}       
\usepackage{microtype}      
\usepackage{xcolor}         

\usepackage{comment}		

\newcommand{\mycite}[1]{\citep{#1}}

\title{Focused Contrastive Training for Test-based Constituency Analysis}

\author{%
  Benjamin Roth, Erion Çano\typeout{\thanks{Use footnote for providing further information
    about author (webpage, alternative address)---\emph{not} for acknowledging
    funding agencies.}} \\
  Digital Philology \\
  Research Group Data Mining and Machine Learning \\
  University of Vienna, Austria \\
  \texttt{first.last@univie.ac.at} \\
}

\begin{document}

\maketitle

\begin{abstract}
We propose a scheme for self-training of grammaticality models for constituency analysis based on linguistic tests.
A pre-trained language model is fine-tuned by contrastive estimation of grammatical sentences from a corpus, and ungrammatical sentences that were perturbed by a \emph{syntactic test}, a transformation that is motivated by constituency theory.
We show that consistent gains can be achieved if only certain positive instances are chosen for training, depending on whether they could be the result of a test transformation.
This way, the positives, and negatives exhibit similar characteristics, which makes the objective more challenging for the language model, and also allows for additional markup that indicates the position of the test application within the sentence.
\end{abstract}

\section{Introduction}
\label{sec:intro}
One way linguists analyze language is by applying \emph{linguistic tests}: Here, transformations that are driven by a specific theory are applied to utterances (e.g., sentences) and if the result of the transformation is judged grammatical (according to the linguist's introspection) then the test has identified the occurrence of a specific category or phenomenon. A prime example of this process is that of \emph{constituency tests}. Constituency theory posits that language is structured hierarchically into constituents, i.e., spans of specific types that function as units and can be moved or replaced by other units of the same type (while other spans that are not constituents do not exhibit those properties). The advantage of test-based linguistic analysis is that it operates very close to the underlying theory, and the principles captured by the tests, stemming from theory, can be very general. In contrast to grammar-based approaches, in a test-based setting, one does not need to spell out all different phenomena that would follow from the theory.

If automated test-based analysis could be successfully implemented, this would theoretically render the laborious manual construction of grammars or training corpora obsolete. However, the test-based linguistic analysis would instead require the correct specifications of tests (transformations) and a grammaticality model (that can replace the linguists' introspection). In this work, we show how to successfully automate a simple class of constituency tests based on \emph{pro-form replacements}. Moreover, we show how a grammaticality model based on annotated training data \typeout{\mycite{warstadt-etal-2019-neural}} can be replaced by one using only unsupervised data. We do this using a proposed mechanism that we refer to as \emph{focused contrastive training}.

Our study reveals the following insights: (i) We compare different strategies for combining the scores of different constituency tests and find that taking the maximum works better than the average or voting. (ii) We systematically compare the performance of different pro-forms, and find a selection of 4 pro-forms (out of 18) that perform markedly better than a previously suggested set. (iii) We show that focused contrastive training outperforms a supervised grammaticality model as well as a previously proposed negative sampling scheme.

\section{Related Work}
\label{sec:relwork}
There is a growing interest in leveraging PLMs (Pretrained Language Models) like BERT and XLNet \mycite{devlin-etal-2019-bert,NEURIPS2019_dc6a7e65} which are based on Transformer self-attention layers \mycite{10.5555/3295222.3295349}. They offer contextual word representations and have proven highly effective in various NLP downstream tasks. Moreover, both PLMs and grammar induction models are trained on the same objective: learn to model natural languages which is useful for several applications. These facts have motivated work to explore the syntactic knowledge captured in PLMs for the possibility of using them in unsupervised constituency parsing: e.g., \citeauthor{marecek-rosa-2019-balustrades} analyze the encoders of Transformer architectures and design a method for extracting constituency trees from self-attention heads. Comparing those trees with standard syntactic parse trees, they conclude that Transformers do indeed capture syntax. Similarly, \citet{li-etal-2020-heads} also extract constituency trees from PLM attention heads. They later rank them based on their properties, revealing multiple insights which could be useful for future research.
 
Constituent parsing with PLMs has been proposed in both supervised \mycite{kitaev-klein-2018-constituency,ijcai2020-560} and unsupervised settings \mycite{DBLP:journals/corr/abs-2004-13805,li-etal-2020-empirical}. Unsupervised constituency parsing is challenging but offers the possibility to work with an unlimited amount of text, overcoming the need for human supervision.  
\citet{DBLP:journals/corr/abs-2004-13805} utilize unsupervised parsing via PLMs by introducing zero-shot consistuency parsing with a chart-based method. It considers every phrase subspan to judge the plausibility of that phrase. They propose an ensemble technique that selects the top-K PLM attention heads and boosts performance.
\citet{cao-etal-2020-unsupervised} propose a semi-supervised approach where they automate a range of movement- and replacement-based constituency tests. Their model is initialized with supervised \mycite{warstadt-etal-2019-neural} and unsupervised grammaticality models
and fine-tuned to optimize the likelihood of an unsupervised chart parser.
In contrast, in our work, we study self-supervised training of grammaticality models that can be used as an isolated modular component in test-based linguistic analysis (independently of parsing algorithms).

\section{Approach}
\label{sec:model}
\subsection{Constituency analysis with linguistic tests}
\label{ssec:constanalysis}

A general approach to linguistic analysis is to reformulate and replace parts in question with prototypical realizations of a phenomenon, and then judge the result with respect to its grammatical acceptability. 
If such reformulations are formalized according to a linguistic theory, standardized and operationalized, they can be called a \emph{linguistic test}.

Given a set of transformations (tests) $T$ each test $t \in T$ is a function that takes a word sequence (utterance or sentence) $s$ as well as a contained subspan $x$ as an input and outputs a transformed word sequence $s^{tx} = t(s,x)$. Moreover, there is a real-valued function $\alpha$ where $\alpha(s')>\alpha (s'')$  for two sentences $s'$ and $s''$ iff $s'$ is more acceptable than $s''$. 

A constituent is any group of words that function as a single unit in a hierarchical structure. 
One main type of constituent tests are \emph{pro-form substitution tests:} Can the constituent candidate be replaced by a pro-form of the same category? The testing procedure is therefore to (1.) substitute the constituent candidate (2.) judge whether the sentence is still acceptable. 
In the following, we study how different choices for automating those two steps influence the quality of automated test-based constituent analysis.

\subsection{Pro-forms}
We created an initial list of pro-forms to cover the most common syntactic categories and include the most common pronouns, including some variability in order to allow for agreement in different contexts. This list also includes the pro-forms suggested by \citet{cao-etal-2020-unsupervised} (underlined). \textbf{Pronouns}: \underline{it}, \underline{ones}, this, that, they, I, we, you; \textbf{Pro-PPs} (preposition+pronoun):  of it, for it, in it; \textbf{Pro-VPs}: \underline{did so}, do that, does that; \textbf{Pro-sentences}: it is, that it is; \textbf{Pro-adverbs}: there, this way.

\subsection{Supervised training}
\label{ssec:supervisedtrain}
A supervised approach for arriving at a grammaticality model $\alpha$ is to train on labeled grammatical and ungrammatical sentences. Similar to \citet{cao-etal-2020-unsupervised}, we train a supervised grammaticality classifier using the CoLA dataset \mycite{warstadt-etal-2019-neural}.

\subsection{Combining scores of different pro-forms} 
\label{ssec:combining_scores}

For a grammaticality model $\alpha$, and input sentences $s$, a set of tests $T$, and a selected subspan $x$, we compare different strategies for combining the scores. \textbf{Maximum}, the highest scoring transformation is used: $\mbox{score}(s,x) = \max_{t \in T} \alpha(t(s,x))$; \textbf{Average}, the average grammaticality score of all tests is used: $\mbox{score}(s,x) = \frac{1}{|T|} \sum_{t \in T} \alpha(t(s,x))$; \textbf{Voting}, two subspans are not directly compared by a score, but by the number tests that score higher for one of them: $|\{t \in T : \alpha(t(s,x)) > \alpha(t(s,x^{'})) \}| > \frac{|T|}{2}$.

\subsection{Contrastive training}
\label{ssec:negsamptrain}

In contrastive training with negative sampling \mycite{8618399,riedel-etal-2013-relation,ma-collins-2018-noise,10.1145/3184558.3186905,10.5555/2999792.2999959}, positive training instances are taken from a set of observations, i.e. drawn from the data distribution, while negative training instances are constructed such that they are unlikely to be generated from the data distribution.

Contrastive training has been used to pre-train a grammaticality model \mycite{warstadt-etal-2019-neural}, using permutations of sentences as corrupted negative samples, based on the assumptions that observed sentences are grammatical and corrupted sentences are ungrammatical.
This insight was incorporated by \citet{cao-etal-2020-unsupervised}, who used a simple form of contrastive \emph{pre}-training for a grammaticality model for constituent tests. 
More precisely, they use corruptions based on applications of constituency tests (applied to randomly selected subspans of observed sentences) as negative samples. Since only a small minority of subspans are constituents, it is likely that the results of these transformations are indeed ungrammatical.\footnote{Assuming binary branching, there are $n\frac{(n-1)}{2}$ subspans but only $n$ constituents of length $\geq 2$. Moreover, not every test is a valid replacement for each constituent type.}
More formally, if $C$ is corpus of observations, then:
\begin{itemize}
\item $X_{pos} \sim C$ are the positive instances, a subset of instances uniformly sampled from $C$ without replacement ($X_{pos} = C$ if the entire corpus is used).
\item $X^{t}_{neg} \sim \{t(s,x): s \in C, x \in \mbox{subspans}(s)\}$ are instances corrupted by a test transformation $t \in T$. In order to achieve an equal ratio between positive and negative samples, we require that $|X_{t}|=\frac{|X_{pos}|}{|T|}$ for all $t \in T$.  The negative instances are then $X_{neg} = \cup_{t \in T} X^{t}_{neg}$ . 
\end{itemize}

We refer to this scheme as \emph{non-focused contrastive training}, and we take all sentences in the training corpus as positive examples $X_{pos}$. We create an equal amount of negative samples, by first uniformly sampling a test (a pro-form) $t \in T$ and a subspan of $2-4$ tokens for each of the (same) sentences, replacing the selected span by the pro-form.
The training or test input to the grammaticality model is the sequence of words without any further markup.
At test-time, the grammaticality score $\alpha(s^{tx})$ will indicate how much the trained model estimates a sentence (after applying a test transformation) to have characteristics of naturally occurring sentences vs. the perturbed ones.

\subsection{Focused contrastive training}
\label{ssec:focnegsamp}

While the setting described above makes sense at the first glance, it is a suboptimal contrastive training scheme in the case of syntactic constituency tests: During training, tests are only applied to the negative samples, hence the negative samples exhibit certain patterns more frequently which stem specifically from the transformations. For this reason, the model might learn to generally give a low grammaticality score whenever a test transformation has been applied. However, at test time there are only transformed sentences, and the transformations of constituent spans should result in a high grammaticality score whenever a constituent has been replaced. Moreover, in the above setting, the grammaticality judgment is about the entire sentence, without any indication about the place that has been affected by the test transformation. However, when applying a test, the question is whether it was the transformation of a specific span that rendered the sentence ungrammatical.

To overcome these weaknesses, we propose \emph{focused contrastive training}:  Our method focuses on \emph{relevant} positive instances that \emph{could be the outcome of a test application}. Since these instances exhibit similar patterns pertaining to specific transformations as the corrupted instance, this makes training more challenging but also avoids problematic reliance on artifacts of training data construction. Similarly, balanced sampling of negative instances explained below, adapts aggregate characteristics of the negative set to those of the positive set. Our method focuses on the part of the sentence that is (or, could have been) affected by a test, providing the relevant information to the model during training and testing. We formalize focused contrastive training:
\begin{itemize}
\item $X^t_{pos} \sim \{s \in C : \exists s',x \mbox{ s.t. } s=t(s',x)\}$ 
are a sample of those instances observed in the corpus that could also have been the result by applying test $t$ to some hypothetical sequence of words $x'$. $X_{pos} = \cup_{t \in T} X^{t}_{pos}$ are the positive instances. 
\item $X^{t}_{neg} \sim \{t(s,x): s \in C, x \in \mbox{subspans}(x)\}$ are instances corrupted by a test transformation $t \in T$, as before. However, since also the positive instances can now be assigned to a specific test, the negative and positive instances can be balanced per test, which avoids test-specific artifacts to be associated with the positive or negative class, by requiring $|X^{t}_{neg}| = |X^{t}_{pos}|$. The negative instances are again $X_{neg} = \cup_{t \in T} X^{t}_{neg}$. 
\end{itemize}
For a more focused representation of each instance in $X^t_{pos}$ and $X^t_{neg}$, we define $s^{tx}= \mbox{markup}(t,s,x)$, where $\mbox{markup}(t,s,x)$ is 
the transformed sentence $t(s,x)$ together with
any additional information about the specific test applied and/or a real or hypothetical sentence as the input to $t$. It is important to note that this markup could not be added for positive instances in the non-focused setting. In our case of pronominalization tests, we simply mark the start and end positions of the pronouns associated with test $t$. The grammaticality model is then used to evaluate $\alpha(s^{tx})$.

Specifically, for \emph{focused contrastive training}, we first create each positive set $X_{pos}^t$ by collecting all sentences in the corpus in which the pro-form of $t$ occurs.
The occurrence of the pro-form is then marked with start (\texttt{<S>}) and end (\texttt{<E>}) markers to create the input for the grammaticality model, for example:

[\emph{Since the last time he traveled <S> this way <E> several months ago , he has recanted a series of bold forecasts of a recession .}]

Then, for each test $t$ the corresponding amount of negative examples $X_{neg}^t$ is created by randomly selecting sentences from the training corpus, and replacing a subspan of $2-4$ tokens with the pro-form of $t$, indicated with start and end markers, for example:

[\emph{In the past year , both have <S> this way <E> the limits of their businesses .}]

\subsection{Data sets}
\subsubsection{Development and test set}

In order to evaluate whether a grammaticality model scores the resulting replacements of constituents higher than that of non-constituents, we create pairs of two subspans $x^{c}, x^{n}$ within the same sentence $s$, where one is a constituent and the other is not.

For every scoring method, we measure its accuracy by the average number of sentences where $\mbox{score}(s,x^{c}) > \mbox{score}(s,x^{n})$. We measure the relative score difference within a sentence in order not to disadvantage scoring models for which the grammaticality score might be strongly impacted by parts of the sentence unrelated to the selected subspan.

Following \citet{chen2014fast,dyer2015transition}, we use sections 22 and 23 of the WSJ portion of the Penn Treebank\footnote{This corpus is licensed via the LDC under catalog number LDC99T42.} as development and test data, respectively. We keep sentences with at least 3 tokens only and remove the shorter ones. For each sentence, we randomly sample a constituent from the treebank annotation (excluding trivial constituents spanning only one token or the entire sentences), as well as a non-constituent span with the corresponding length. This results in 1,672 sentences for the development set and 2,348 sentences for the test set.

\subsubsection{Training sets and RoBERTa model}
\label{sssec:traindetails}
\textbf{CoLA.} For supervised training, we use the grammaticality annotations of 10,657  sentences from \citet{warstadt-etal-2019-neural}.\\
\textbf{WSJ.} For contrastive training, we use sections 02-21 of the WSJ portion of the Penn Treebank which total in 37,374 sentences. We do not make use of any constituent annotation, as our goal is the unsupervised training of the grammaticality model.\\
\textbf{RoBERTa.} For all experiments, we fine-tune a pre-trained RoBERTa-base \mycite{liu2019roberta} model from the huggingface transformer library,\footnote{\url{https://huggingface.co/transformers/}} following the hyper-parameter choices of \citet{liu2019roberta,cao-etal-2020-unsupervised} for the CoLA task. We train for 10 epochs, and choose the model with the best accuracy on the dev data.

Our code is written in Python version 3.8.10. We ran our experiments on a DGX-1 server with Ubuntu 20.04 GNU/Linux using one Nvidia V100 GPU per experiment. The code execution of each experiment took 2h or less. 

\begin{table}
\centering
\begin{tabular}{l c}
\hline
pronoun set & accuracy \\ [0.17ex]
\hline
this way & 0.7500 \\ [0.17ex]
this way, did so & 0.7763 \\ [0.17ex]
this way, did so, of it & 0.7883 \\ [0.17ex]
this way, did so, of it, it & 0.7996 \\ [0.17ex]
\hline 
it, ones, did so & 0.6848 \\ [0.17ex]
\hline
\end{tabular} 
\caption{Accuracy (dev set) for selected pronoun sets, using the CoLA pre-trained model. Greedy selection according to the \emph{maximum} selection strategy (top), and pronoun set of Cao et al. (bottom).}
\label{tab:pronoun_sets}
\end{table}

\section{Results and Discussion}
\label{sec:results}

\begin{table}[t]
\centering
\begin{tabular}{l l c c c}
\hline
data & scheme & train-full & train-greedy & train-complement \\ [0.17ex]
\hline
dev & non-focused & 0.8188 & 0.7075 & 0.7913 \\ [0.17ex]
test & non-focused & 0.8399 & 0.7189 & 0.7841 \\ [0.17ex]
\hline
dev & focused & 0.8433 & 0.8242 & 0.8206 \\ [0.17ex]
test & focused & 0.8560 & 0.8152 & 0.8160 \\ [0.17ex]
\hline
\end{tabular} 
\caption{Effect of focused contrastive training using different pronoun sets during training (the \emph{greedy} pronoun set is used for scoring).
}
\label{tab:pn_comparison_training}
\end{table}

In a first experiment, we investigated the performance of different score combination schemes (\emph{maximum}, \emph{average} and \emph{voting}, see Section~\ref{ssec:combining_scores}). 
For this, we used the grammaticality model trained with the supervised CoLA dataset and got the following accuracy scores for three strategies on the dev set: \emph{maximum}: 0.7697, \emph{average}: 0.756, \emph{voting}: 0.7075.   
Hence, the most effective strategy for combining different tests is taking the \emph{maximum} score. This is intuitive because for detecting a constituent it should suffice that one of the tests is appropriate, and it is to be expected that different tests work well in different contexts. The \emph{maximum} strategy is therefore used for all other experiments.  

The results in Table~\ref{tab:pronoun_sets} show that strong improvements can be achieved by selecting a subset of pronouns and that the subset of pronouns we arrived at performs more than $10\,\%$ accuracy points better than the one which was originally proposed by \citet{cao-etal-2020-unsupervised}. 
Table~\ref{tab:examples} shows some example sentences with selected subspans, together with the pronouns that yielded the highest score when substituted.

Tables \ref{tab:pn_comparison_training} and \ref{tab:pn_comparison_application} show that unsupervised contrastive training generally yields comparable or better acceptablity estimates for constituency analysis than the CoLA-based model.
Experiments with different sets of pronouns for training and test application reveal the robustness of our approach:
In particular, pretraining with a set of pronouns that is complementary to those pronouns used for constituency analysis yields good results as well.
In general, training with a large set of pronouns and performing constituent analysis with a selection of pronouns works best.
Focused contrastive training, i.e. selecting and marking positive and negative samples so that they are most helpful for decisions at analysis time has a consistently positive effect in all settings.

\begin{table}[h]
\centering
\begin{tabular}{l l c c c}
\hline
data & scheme & score-full & score-greedy & Cao et al. \\ [0.17ex]
\hline
dev & non-focused & 0.7075 & 0.8188 & 0.7243  \\ [0.17ex]
test & non-focused & 0.7142 & 0.8399 & 0.7470 \\ [0.17ex]
\hline
dev & focused & 0.7907 & 0.8433 & 0.7518 \\ [0.17ex]
test & focused & 0.7922 & 0.8560 & 0.7602 \\ [0.17ex]
\hline
\end{tabular} 
\caption{Comparison of different pronouns sets used for scoring (for the model trained on the \emph{full} pronoun set).
}
\label{tab:pn_comparison_application}
\end{table}

\begin{table}[h] 
\onecolumn
\centering
\begin{tabular}{l p{0.76\textwidth} l l}
\hline
& sentence & pred.? & const.? \\ [0.17ex]
\hline
%
\emph{orig.} & Mr. Sim figures it will be easier to turn Barry Wright around since he 's now in the driver 's seat . &  &  \\ [0.17ex]
\emph{repl.} & Mr. Sim figures it will be easier to turn Barry Wright around \textbf{this way} . & Yes & Yes \\ [0.17ex]
\emph{repl.} & Mr. Sim figures it will be easier to turn Barry \textbf{it} 's seat . & No & No \\ [0.17ex]
\hline
\emph{orig.} & Other fund managers were similarly sanguine . &  &  \\ [0.17ex]
\emph{repl.} & Other fund managers \textbf{did so} . & Yes & Yes \\ [0.17ex]
\emph{repl.} & Other fund \textbf{of it} sanguine . & No & No \\ [0.17ex]
\hline
\emph{orig.} &  On Saturday night , quite a few of the boys in green and gold salted  &  &  \\ [0.17ex]
& away successes to salve the pain of past and , no doubt , future droughts . &  &  \\ [0.17ex]
\emph{repl.} &  On Saturday night , quite a few of the boys in green and gold salted  away successes to salve \textbf{did so} . & No & Yes \\ [0.17ex]
\emph{repl.} &  \textbf{This way} green and gold salted away successes to salve the pain  of past and , no doubt , future droughts . & Yes & No \\ [0.17ex]
%
%
\hline
\end{tabular} 
\caption{\label{tab:examples}
Example sentences from the dev data. Given is the original sentence, as well as transformed versions from replacing a selected subspan with the pronoun scored highest (by the CoLA model). The last columns indicate whether the replaced span was predicted to be a constituent, and whether it actually was one (according to the treebank).}
\end{table}

\bibliography{neurips_2021}
\bibliographystyle{acl_natbib}

\end{document}